\documentclass[10pt,twocolumn,letterpaper]{article} 
\usepackage[submission]{aaai23}  
\usepackage{times}  
\usepackage{helvet}  
\usepackage{courier}  
\usepackage[hyphens]{url}  
\usepackage{graphicx} 
\usepackage{natbib}  
\usepackage{caption} 
\usepackage{algorithm}
\usepackage{amsmath,amsfonts,bm}
\usepackage{amssymb}
\usepackage{algpseudocode}
\usepackage{dsfont}
\usepackage{newfloat}
\usepackage{listings}
\DeclareCaptionStyle{ruled}{labelfont=normalfont,labelsep=colon,strut=off} 
\floatstyle{ruled}
\newfloat{listing}{tb}{lst}{}
\floatname{listing}{Listing}
\newcommand{\Xset}{\mathcal{X}}

\newcommand{\Yset}{\mathcal{Y}}

\begin{document}

\author{
    Xinran Liu\textsuperscript{\rm 1}, 
    Yikun Bai\textsuperscript{\rm 1},
    Yuzhe Lu\textsuperscript{\rm 1},
    Andrea Soltoggio\textsuperscript{\rm 2},
    Soheil Kolouri\textsuperscript{\rm 1}
}
\affiliations{
    \textsuperscript{\rm 1}Department of Computer Science, Vanderbilt University\\
    \textsuperscript{\rm 2}Department of Computer Science, Loughborough University
}
\title{Wasserstein Task Embedding for Measuring Task Similarities}
\maketitle

\begin{abstract}
Measuring similarities between different tasks is critical in a broad spectrum of machine learning problems, including transfer, multi-task, continual, and meta-learning. Most current approaches to measuring task similarities are architecture-dependent: 1) relying on pre-trained models, or 2) training networks on tasks and using forward transfer as a proxy for task similarity. In this paper, we leverage the optimal transport theory and define a novel task embedding for supervised classification that is model-agnostic, training-free, and capable of handling (partially) disjoint label sets. In short, given a dataset with ground-truth labels, we perform a label embedding through multi-dimensional scaling and concatenate dataset samples with their corresponding label embeddings. Then, we define the distance between two datasets as the 2-Wasserstein distance between their updated samples. Lastly, we leverage the 2-Wasserstein embedding framework to embed tasks into a vector space in which the Euclidean distance between the embedded points approximates the proposed 2-Wasserstein distance between tasks. We show that the proposed embedding leads to a significantly faster comparison of tasks compared to related approaches like the Optimal Transport Dataset Distance (OTDD). Furthermore, we demonstrate the effectiveness of our proposed embedding through various numerical experiments and show statistically significant correlations between our proposed distance and the forward and backward transfer between tasks on a wide variety of image recognition datasets. 
\end{abstract}
\section{Introduction}

Learning from a broad spectrum of tasks and transferring knowledge between them is a cornerstone of intelligence, and primates perfectly exemplify this characteristic. Modern Machine Learning (ML) is rapidly moving toward multi-task learning, and there is great interest in methods that can integrate, rapidly adapt, and seamlessly transfer knowledge between tasks. When learning from multiple possibly heterogeneous tasks, it is essential to understand the relationships between the tasks and their fundamental properties. It is, therefore, highly desirable to define (dis)similarity measures between tasks that will allow one to cluster tasks, have better control over the forward and backward transfer, and ultimately require less supervision for learning tasks.  

There has been an increasing interest in assessing task similarities and their relationship with forward and backward knowledge transfer among tasks. For instance,  various recent works look into the selection of good source tasks/models for a given target task to maximize the forward transfer to the target task \cite{achille2019task2vec,zamir2018taskonomy,bao2019information,bhattacharjee2020p2l,fifty2021efficiently}. Others have demonstrated the relationship between negative backward transfer (i.e., catastrophic forgetting) and task similarities \cite{nguyen2019toward}. 

Many existing methods for measuring task similarities depend on the choice of model(s), architecture(s), and the training process \cite{leite2005predicting,zamir2018taskonomy,achille2019task2vec,khodak2019adaptive,nguyen2020leep, venkitaraman2020task,gao2021information}. For example, \citet{zamir2018taskonomy,venkitaraman2020task,gao2021information} use pre-trained task specified models to measure a notion of forward transfer and define it as task similarity. \citet{achille2019task2vec} embed tasks into a vector space that relies on a partially-trained network.  \citet{khodak2019adaptive} use the optimal parameters as a proxy for each task and \citet{leite2005predicting} use the learning curves of a pre-specified model to measure task similarities. Besides being model-dependent, these approaches are often computationally expensive as they involve training deep models (or require pre-trained models). 

Model-agnostic task similarity measures provide a fundamentally different approach to quantifying task relationships \cite{ben2006analysis,alvarez2020geometric,tran2019transferability,tan2021otce}. These methods often measure the similarity between tasks as a function of the similarity between the joint or conditional input/output distributions, sometimes also taking the loss function into account. The classic theoretical results for such similarity measures \cite{ben2006analysis,batu2000testing} focus on information theoretic divergences between the source and target distributions. More recently, Optimal Transport (OT) based approaches \cite{alvarez2020geometric,tan2021otce,xu2022selecting} have shown promise in modeling task similarities. Notably, \citet{alvarez2020geometric} approach measuring task similarities through the lens of a hierarchical OT \cite{yurochkin2019hierarchical} where they solve an inner OT problem to calculate the label distance between the class-conditional distributions of two supervised learning tasks. The label distance is then incorporated into the transportation cost of an outer OT problem, resulting in a distance between two datasets that integrates both sample and label discrepancies. \citet{tan2021otce} treats the optimal transport plan between the input distributions of two tasks as a joint probability distribution and use conditional entropy to measure the difference between the two tasks. One major shortcoming of these OT-based approaches is their computational complexity. These methods require the pairwise calculation of OT (or entropy regularized OT) between different tasks, which can be prohibitively expensive in applications requiring frequent evaluations of task similarities, e.g., in continual learning. 

We propose a novel OT-based task embedding for supervised learning problems that is model-agnostic and computationally efficient. On the one hand, our proposed approach is similar to \citep{achille2019task2vec} and \citep{peng2020domain2vec}, which embed datasets into a vector space in which one can easily measure the difference between tasks, e.g., via the Euclidean distance between embedded vectors. On the other hand, our approach is inspired by the Optimal Transport Dataset Distance (OTDD) \cite{alvarez2020geometric} framework, and it essentially provides a Euclidean embedding for a hierarchical OT-based distance between tasks. To calculate such a task embedding, we use the Wasserstein embedding framework \cite{wang2013linear,kolouri2020wasserstein}. Importantly, our approach alleviates the need for pairwise calculation of OT problems between tasks, turning it into a more desirable solution than previously proposed methods. 

{\bf Contributions.} We propose a computationally efficient and model-agnostic task embedding, denoted as Wasserstein Task Embedding (WTE), in which the Euclidean distance between embedded vectors approximates a hierarchical OT distance between the tasks. We provide extensive numerical experiments and demonstrate that: 1) our calculated distances between embedded tasks are highly correlated with the OTDD distance \cite{alvarez2020geometric}, 2) our proposed embedding and similarity calculation is significantly faster than the OTDD distance, and 3) our proposed similarity measure provides strong and statistically significant correlation with both forward and backward transfer.

\section{Related work}
\textbf{Model-based task similarity}. Most existing approaches to measuring task similarity are model-dependent and use forward transferability as a proxy for similarity. 
For example, \citet{zamir2018taskonomy} use pre-trained models on source tasks and measure their performance on a target task to obtain an asymmetric notion of similarity between source and target tasks.
Following \citet{zamir2018taskonomy}'s work, \citet{dwivedi2019representation} measure the transferability in a more efficient manner by applying the Representation Similarity Analysis (RSA) between the trained models (e.g. DNNs) from different tasks. 
Similarly, \citet{nguyen2020leep} assume the source and target tasks share the same set of inputs but have different sets of labels, and estimate the transferability by the empirical conditional distribution of target labels given the inputs computed by a pre-trained model on the source task. 

Another class of approaches embed the tasks into a vector space and then define the (dis)similarity on the embedded vector representations. \citet{achille2019task2vec} discuss processing data (images) through a partially trained ``probe network'' and obtain vector embedding by computing the Fisher information matrix (FIM). The (dis)similarity of two tasks is then computed from the difference between the the FIMs. Similarly, \citet{peng2020domain2vec} propose a domain (labeled dataset) to vector technique. In particular, given a domain, they feed the data to a pre-trained CNN to compute the Gram matrices of the activations of the hidden convolutional layers, and apply feature disentanglement to extract the domain-specified features. Concatenation of the diagonal entries of Gram matrices and the domain-specified features gives the final domain embedding.
These methods, however, highly rely on the pre-trained models and training process, and lack theoretical guarantees. On the opposite side of the spectrum is directly measuring the discrepancy between domains.

\textbf{Discrepancy measures of domains}. 
Over the years, numerous notions of discrepancy to measure the (dis)similarity of datasets (domains) were proposed, including $L_1$-norm \citep{batu2000testing}, generalized Kolmogorov-Smirnov distance \citep{devroye1996parametric,kifer2004detecting},
and loss-oriented discrepancy distance \citep{mansour2009domain}. In the context of domain adaptation, generalized Kolmogorov-Smirnov distance (later known as the $\mathcal{A}$-distance) is a principled notation of discrepancy, which is a relaxation of total variation. \citet{ben2006analysis} show that the target performance (generalization error) is controlled by the empirical estimate of the source domain error and the $\mathcal{A}$-distance between source and target domains. 
Another widely used distance is the Maximum Mean Discrepancy (MMD) \cite{gretton2006kernel}, which captures the (dis)similarity of the  embedding of distribution measures in a reproducing kernel Hilbert space. \citet{pan2010domain} propose to learn transfer components across domains in reproducing kernel Hilbert space using MMD, and show that the subspace spanned by these transfer components preserves data properties. Such domain discrepancy methods, however, can not take labels into account, and thus may not be enough to reflect the similarity of tasks. 



\textbf{Optimal transport based task similarity}.
In recent years, metrics rooted in the optimal transport problem, e.g., the ``Wasserstein distance'' \cite{villani2009optimal,villani2021topics} (or the ``earth mover's distance'' \cite{rubner2000earth,solomon2014earth}), have attracted growing interest in the machine learning community. Wasserstein distance is a rigorous metric of probability measures endowed with desired statistical convergence behavior, in contrast to other classical discrepancies (e.g. KL-divergence, total variation, JS-divergence, Hellinger distance, Maximum mean discrepancy, etc). OT based metrics are widely used in generative modeling \citep{arjovsky2017wasserstein,liu2019wasserstein}, domain adaptation \citep{courty2017joint,courty2014domain,alvarez2018gromov}, graph embedding \cite{kolouri2020wasserstein,xu2019gromov}, and neural architecture search \citep{kandasamy2018neural}.

\citet{alvarez2020geometric} propose a notion of distance between two datasets in a supervised learning setting. They introduce Optimal Transport Dataset Distance (OTDD) based on the OT theory, which can be thought as a hierarchical OT distance where the transportation cost measures the distance between samples as well as labels. With the assumption that the label-induced distributions can be approximated by Gaussians, the distance between labels is defined as the Bures-Wasserstein distance.

\citet{tan2021otce} introduce another OT-based method to measure the transferability, named OTCE (Optimal Transport Conditional Entropy) score. In particular, they first use the entropic optimal transport to estimate domain differences and then use the optimal coupling between the source and target distributions to compute the conditional entropy of the target task given source task. The OTCE is defined by the linear combination of the OT distance and the conditional entropy.   
Both OTDD and OTCE were shown to be effectively aligned with forward transfer, however, the computation of the pairwise Wasserstein distances among increasing number of datasets remains expensive. This hinders the application of these methods to problems where one needs to perform nearest dataset retrieval frequently (e.g., memory replay approaches in continual learning).


\textbf{Computation Cost of OT Distance}. Calculating the Wasserstein distance involves solving an $n^2$ dimension linear programming and the computational cost is $\mathcal{O}(n^3\log(n))$ for a pair of $n$-size empirical distributions \citep{pele2009fast}. To facilitate the computation, one common method is adding entropic regularization \citep{cuturi2013sinkhorn,peyre2017computational}, by which the original linear programming problem is converted into a strictly convex problem. By applying the Sinkhorn-Knopp algorithm \citep{peyre2017computational,chizat2018scaling} to find an $\epsilon$- accurate solution, the computational complexity reduces to $\mathcal{O}( n^2\log(n)/\epsilon^3)$ \citep{altschuler2017near}. However, this technique suffers a stability-accuracy trade-off. When the regularity coefficient is high, the objective is biased toward the entropy term; when it is small, the Sinkhorn algorithm will not be numerically stable. 
\begin{figure*}[t!]
    \centering
    \includegraphics[width=0.85\linewidth]{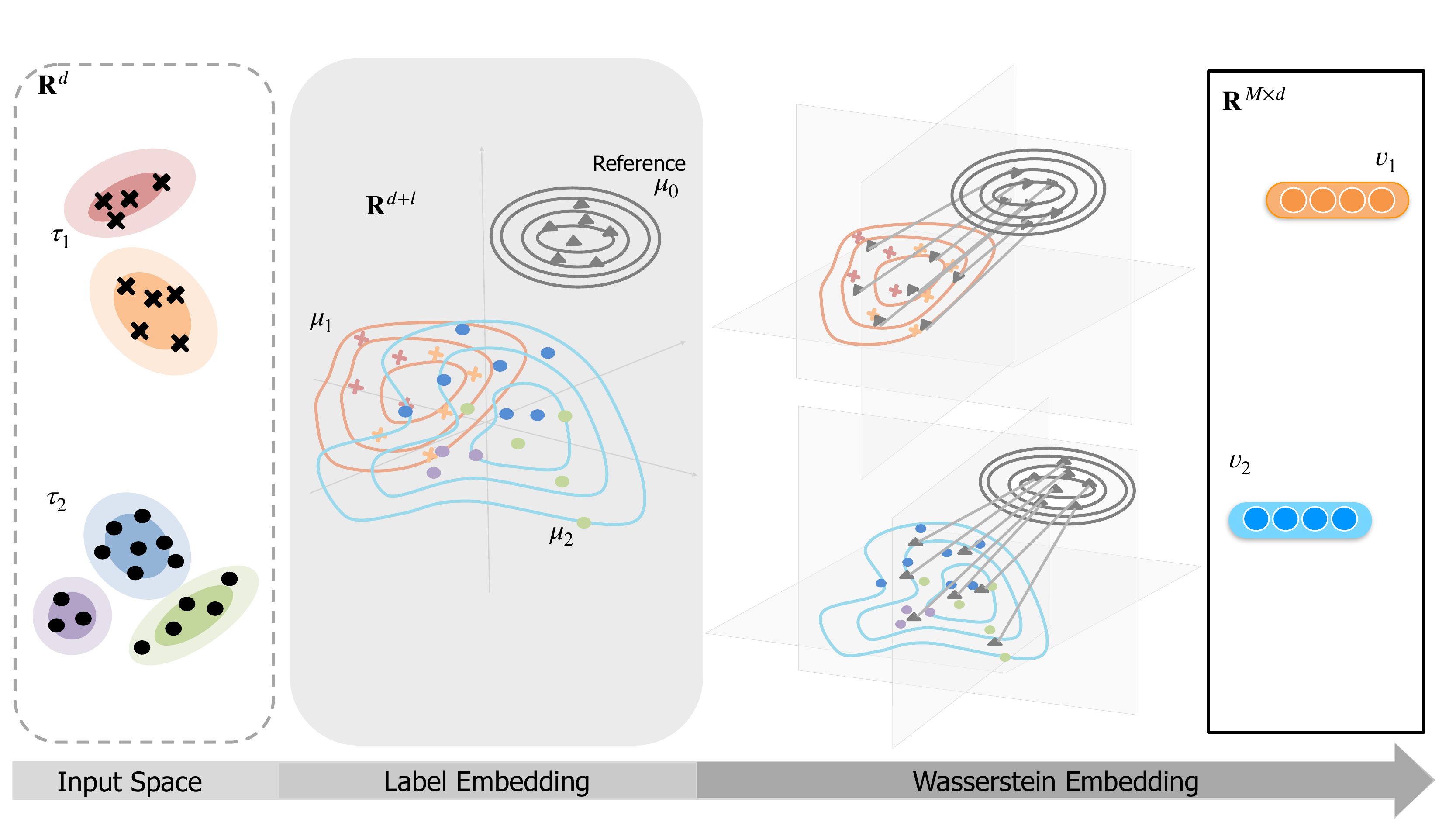}
    \caption{Wasserstein Task Embedding framework. Given tasks $\tau_1$ and $\tau_2$ with input space $\mathbb{R}^d$, WTE first map them into $\mathbb{R}^{d+L}$ as probability distributions $\mu_1$ and $\mu_2$ by MDS, then apply WE to get vector $v_1$ and $v_2$ with respect to a fixed reference measure $\mu_0$. Here $M$ is the size of reference set.}
    \label{fig:framework}
\end{figure*}


\section{Preliminaries}
\subsection{Multidimensional scaling (MDS)} 
Multidimensional scaling (MDS) \cite{cox2008multidimensional} is a non-linear dimensionality reduction approach that embeds $N$ samples into an $l$-dimensional Euclidean space while preserving their pairwise distances. Given a set of high-dimensional data $\Xset=\{x_n\}_{n=1}^N$ and the proximity matrix $D\in \mathbb{R}^{N\times N}$, where $D_{i,j} = d(x_i, x_j)$, and $d(\cdot, \cdot)$ denotes the metric in $\Xset$, the goal of MDS is to construct a distance-preserving map from $\Xset$ to a lower-dimensional Euclidean space $\mathbb{R}^{l}$. Depending on the objective and inputs, MDS can be classified into metric MDS and non-metric MDS. Specifically, metric MDS aims to find a map $\psi:\Xset\rightarrow\mathbb{R}^{l}$ such that
\begin{align}
\label{eq:mds_def}
    \min_{\psi}\sqrt{\frac{\sum_{i,j}\left(d(x_i, x_j)-\|\psi(x_i), \psi(x_j)\|\right)^2}{\sum_{i,j}d(x_i, x_j)^2}},
\end{align}
which can be solved by Algo. \ref{algo:mds}.
\begin{algorithm}
    \caption{Multidimensional Scaling}
    \label{algo:mds}
    \begin{algorithmic}
    \Procedure{MDS}{$\Xset=\{x_n\}_{n=1}^N,~D=[d(x_i,x_j)]_{i,j},~l$}
    \State $B=-\frac{1}{2}(id-\frac{1}{N}\mathds{1}_{N})D(id-\frac{1}{N}\mathds{1}_{N})$
    \State Eigen-decomposition $B=V\Lambda V^T$
    \State Rearrange $\Lambda$ into $\hat{\Lambda}$ with descending order of variances
    \State Rearrange $V$ into $\hat{V}$ in correspondence with $\hat{\Lambda}$
    \State $\hat{\Lambda}_{(l)}=\hat{\Lambda}[:l,:l]$; $\hat{V}_{(l)} = \hat{V}[:l]$
    
    \Return $\psi(\Xset)=\hat{\Lambda}_{(l)}^{\frac{1}{2}}\hat{V}_{(l)}$ 
    \EndProcedure
    \end{algorithmic}
\end{algorithm}
Note that MDS not only works for Euclidean distances, but also for other dissimilarities such as Wasserstein distances \cite{5609205, hamm2022wassmap}.
\subsection{Wasserstein Distances}
Let $\mu, \nu$ be Borel probability measures on $\Xset\subseteq\mathbb{R}^d$ with finite $p^\text{th}$ moment, and the corresponding probability density functions are $p_\mu$ and $p_\nu$, i.e. $d\mu = p_\mu dx$, $d\nu = p_\nu dx$. The 2-Wasserstein distance between $\mu$ and $\nu$ is defined as \cite{villani2009optimal}:
\begin{equation}
    \label{eq:wd_kantorovich}
    \mathcal{W}_2(\mu, \nu) = \left(\inf_{\gamma\in\Gamma(\mu,\nu)}\int_{\Xset\times\Xset} \|x-x'\|^2 d\gamma(x, x') \right)^{\frac{1}{2}},
\end{equation}
where $\Gamma(\mu, \nu)$ is the set of all transport plans between $\mu$ and $\nu$, i.e. probability measures on $\Xset\times\Xset$ with marginals $\mu$ and $\nu$. We also note that by Brenier theorem \cite{Brenier1991PolarFA}, given two absolutely continuous probability measures $\mu, \nu$ on $\mathbb{R}^d$ with densities $p_\mu, p_\nu$, there exists a convex function $\phi$ such that $T = \nabla\phi$ is a transport map sending $\mu$ to $\nu$. Moreover, it is the optimal map in the Monge-Kantorovitch optimal transport problem with quadratic cost:
\begin{equation}
    \label{eq:wd_Monge}
    \mathcal{W}_2(\mu,\nu) = \left(\int_\Xset \|x-T(x)\|^2 p_\mu dx \right)^{\frac{1}{2}},
\end{equation}
where $T=\nabla\phi$ pushes $\mu$ to $\nu$, denoted by $T_\#\mu=\nu$.
\subsection{Wasserstein Embedding (WE)}
Wasserstein Embedding \citep{wang2013linear,kolouri2016continuous, courty2017learning, kolouri2021wasserstein} provides a Hilbertian embedding for probability distributions such that the Euclidean distance between the embedded vectors approximates the 2-Wasserstein distance between the two distributions. Let  $\{\mu_i\}_{i=0}^I$ be a set of a probability distributions over $\Xset\subseteq\mathbb{R}^d$ with densities $\{p_i\}_{i=0}^I$. We fix $\mu_0$ as the reference measure. Assume $T_i$ is the optimal transport map that pushes $\mu_0$ to $\mu_i$, the Wasserstein embedding of $\mu_i$ is through a function $\Phi$ defined as
\begin{equation}
    \label{eq:linear we}
    \Phi(\mu_i) = (T_i-id)\sqrt{p_0}
\end{equation}
where the $id$ is the identity function, i.e., $id(x)=x$. $\Phi$ admits nice properties including but not limited to \cite{kolouri2021wasserstein}:
\begin{enumerate}
    \item $d(\mu_i, \mu_j)\triangleq\|\Phi(\mu_i)-\Phi(\mu_j)\|_2$ is a true metric between $\mu_i$ and $\mu_j$, moreover, it approximates the 2-Wasserstein distance: $d(\mu_i,\mu_j)\approx\mathcal{W}_2(\mu_i,\mu_j)$.
    \item In particular, $\|\Phi(\mu_i)\|_2=\|\Phi(\mu_i)-\Phi(\mu_0)\|_2\approx\mathcal{W}_2(\mu_i, \mu_0)$. Here we leveraged the fact $\Phi(\mu_0)=0$.
\end{enumerate}
Although these hold true for both continuous and discrete measures $\{\mu_i\}_{i=0}^I$, we focus on the (uniformly distributed) discrete setting in this paper and provide the following numerical computation details.
Let $p_i = \frac{1}{N_i}\sum_{n=1}^{N_i}\delta_{x_n^i}$, where $\delta_x$ is the Dirac delta function centered at $x\in\Xset$ and $X_i=\{x_n^i\}_{n=1}^{N_i}$ is the set of locations of non-negative mass for $\mu_i$. Then the Kantorovich problem with quadratic cost between $\mu_i$ and $\mu_0$ can be formulated as 
\begin{equation}
    \label{eq:discrete_ot}
    \min_{\pi\in\Pi_i}\sum_{n=1}^{N_0}\sum_{k=1}^{N_i}\pi_{nk}\|x_n^i-x_k^0\|_2^2
\end{equation}
where the feasible set is 
\begin{equation}
    \label{eq:feasible_set}
    \Pi_i\triangleq\{\pi\in\mathbb{R}^{N_0\times N_i}|N_0\sum_{n=1}^{N_i}\pi_{nk}=N_i\sum_{k=1}^{N_0}\pi_{nk}=1,~\forall n,k\}.
\end{equation}
The optimal transport plan $\pi_i^*$ is the minimizer of the above optimization problem, which is solved by linear program at cost $\mathcal{O}(N^3\log(N))$, $N$ being the number of input samples. To avoid mass splitting, the barycentric projection \cite{wang2013linear} assigns each $x_j^0$ in the reference distribution to the center of mass it is sent to and thus outputs an approximated Monge map $T_i$. Then the Wasserstein Embedding for input $X_i$ is calculated by 
\begin{equation}
    \label{eq:disrete_we}
    \Phi(X_i) =  (T_i-X_0)/\sqrt{N_0}\in\mathbb{R}^{N_0\times d}.
\end{equation}
One of the motivations behind Wasserstein embedding is to reduce the need for computing pairwise Wasserstein distances. Given $M$ datasets, computation of $\frac{M(M-1)}{2}$ Wasserstein distances across all distinct pairs is impractically expensive especially when $M$ is large, while leveraging Wasserstein embedding, it suffices to calculate only $M$ Wasserstein distances and the pairwise Euclidean distances between the embedded distributions.

\section{Method}
In this section, we specify the problem setting, review the OTDD framework, and then propose our Wasserstein task embedding (WTE).
\subsection{Problem Setting}
In supervised classification problems, tasks are represented by input-label pairs and can be denoted as $\tau = \mathcal{X}\times\mathcal{Y} = \{(x_n, y_n)\}_{n=1}^N$, where $\mathcal{X}\subseteq\mathbb{R}^d$ is the data/inputs and $\mathcal{Y}$ is the labels. We aim to define a similarity/dissimilarity measure for tasks that enables task clustering and allows for better control over the forward and backward transfer.
\begin{figure*}[t]
    \centering
    \includegraphics[width=0.85\linewidth]{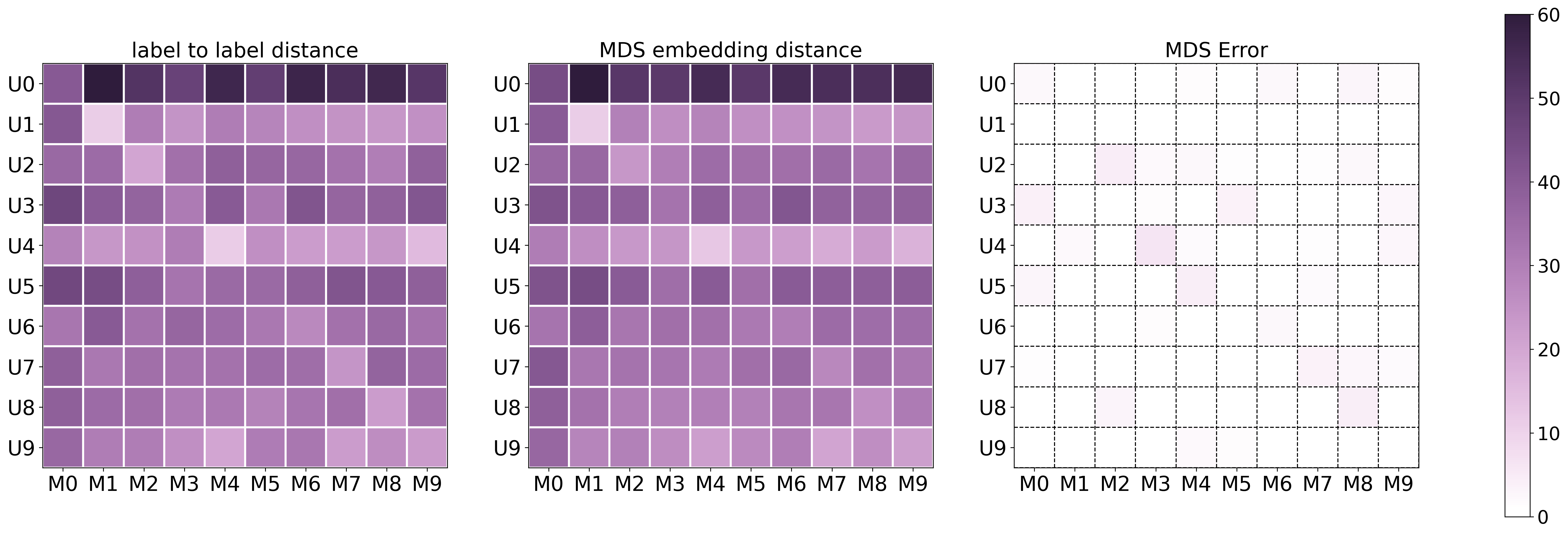}
    \caption{Label-to-label Bures-Wasserstein distance (left) and label MDS embedding Euclidean distances (middle) between MNIST and USPS datasets, squared error is provided on the right.}
    \label{fig:mds_error}
\end{figure*}

\subsection{Optimal Transport Dataset Distance (OTDD)}



Let $\Xset=\{x_n\in \mathbb{R}^d\}_{n=1}^N$ be the input set with labels (classes) $\Yset=\{y_j\}_{j=1}^J$. For each $j$, let $C_{y_j}\subseteq\Xset$ denote the class with label $y_j$. Following the OTDD framework, let $\tau=\{(x_i, y_j)\in \Xset\times\Yset~|~x_i\in C_{y_j} \}_{i,j}$ denote the set of data-label pairs. OTDD encodes the label $y_j$ as distribution $\nu_{y_j}$, where $\nu_{y_j}=\frac{1}{|C_{y_j}|}\sum_{x_i\in C_{y_j}}\delta_{x_i}$. The ground distance in $\tau$ is then defined by combining the Euclidean distance between the data points and the 2-Wasserstein distance between label distributions:
\begin{align}
    d_{\tau}\left((x,y),(x',y')\right):=\left(\|x-x'\|^2+W_2^2(\nu_{y},\nu_{y'})\right)^{\frac{1}{2}}.
    \label{eq:otdd_dist}
\end{align}
Based on this metric, the OT distance between two distributions $\mu_i$ and $\mu_j$ on $\tau$ is
\begin{align}
    d_{OT}(\mu_{i},\mu_{j})=\inf_{\pi\in \Pi(\mu_{i},\mu_{j})}\int_{\tau\times\tau}d_{\tau}(z,z')^2d\pi(z,z'),
    \label{eq:otdd}
\end{align}
where $\Pi(\mu_{i},\mu_{j})$ denotes the set of transport plans between $\mu_{i}$ and $\mu_{j}$. Note that Eq. \ref{eq:otdd} is a hierarchical transport problem, as the transportation cost itself depends on calculation of the Wasserstein distance. To avoid the computational cost of a hierarchical optimal transport problem, \citet{alvarez2020geometric} replace the Wasserstein distance in Eq. \ref{eq:otdd_dist} with the Bures-Wasserstein distance \citep{malago2018wasserstein,bhatia2019bures}, which assumes that $\nu_y$s are Gaussian distributions. Throughout the paper, we consider only the exact-OTDD, as opposed to the entropy-regularized and other variants.

\subsection{Wasserstein Task Embedding}
We define a task-2-vec framework (Fig. \ref{fig:framework}) using Wasserstein embedding (WE) such that the (squared) Euclidean distance between two vectors approximates the OTDD between the original tasks, and denote this embedding by WTE. We later show in the experiment section that the Euclidean distance between the embedded task vectors is not only highly predictive of forward transferability, but also significantly correlates with the backward transferability (catastrophic forgetting). 


\textit{Label Embedding via MDS}. The combination of optimal transport metric with MDS technique was first introduced as an approach to characterize and contrast the distribution of nuclear structure in different tissue classes (normal, benign, cancer, etc.) \cite{5609205}, and further studied in image manifold learning \cite{hamm2022wassmap}. In short, it seeks to isometrically map probability distributions to vectors in relatively low-dimensional space. We leverage the prior work and define an approximated isometry on the label distributions by 1) calculating the pairwise Wasserstein distances and 2) applying MDS to obtain embedded vectors. We adopt the same simplification as in OTDD, that is, assuming the label distributions are Gaussians to replace Wasserstein distances with the closed form Bures-Wasserstein distance:
\begin{equation}
    \label{eq:bures}
    \mathcal{W}_2^2(\nu_y,\nu_{y'})=\|u_y-u_{y'}\|_2^2+\text{Tr}(\Sigma_y+\Sigma_{y'}-2(\Sigma_y^{\frac{1}{2}}\Sigma_{y'}\Sigma_y^{\frac{1}{2}})^{\frac{1}{2}})
\end{equation}
where $u$ and $\Sigma$ denote the mean and covariance matrix of Gaussian distributions. In consistence with previous notations, let us denote the label (MDS) embedding operator by $\psi$, then
\begin{equation}
    \label{eq:mds}
    \mathcal{W}_2^2(\nu_y,\nu_{y'})\approx\|\psi(\nu_y)-\psi(\nu_{y'})\|_2^2,
\end{equation}
where $\psi(\nu_y), \psi(\nu_{y'})\in \mathbb{R}^l$ are vectors whose dimension $l$ is selected to balance the trade-off between accuracy and computation cost \cite{tenenbaum2000global}. Having both inputs and labels represented as vectors, we concatenate these two components and map the data-label pairs $\tau$ to $\tau'\subseteq\mathbb{R}^{d+l}$ such that 
\begin{align}
    \label{eq:lbl_emb}
    d_{\tau}\left((x,y),(x',y')\right)&\approx (\|x-x'\|_2^2+\|\psi(\nu_y)-\psi(\nu_{y'})\|_2^2)^\frac{1}{2}\nonumber\\
    &=(\|[x,\psi(\nu_y)]-[x',\psi(\nu_{y'})]\|_2^2)^\frac{1}{2}\nonumber\\
    &=\|[x,\psi(\nu_y)]-[x',\psi(\nu_{y'})]\|_2
\end{align}
where $[\cdot,\cdot]$ denotes the concatenation operator and the domain $\tau'=\{[x,\psi(\nu_y)]\}_{x\in\Xset}$ is equipped with $l_2$ norm. 
Fig. \ref{fig:mds_error} shows this approximation performance among labels in MNIST  \cite{LeCun2005TheMD} and USPS \cite{291440} datasets. MDS embeddings can capture the pairwise relationships with a maximum of $7.26\%$ error by $10$-dimensional vectors.
\begin{figure*}[t!]
    \centering
    \includegraphics[width=0.9\linewidth]{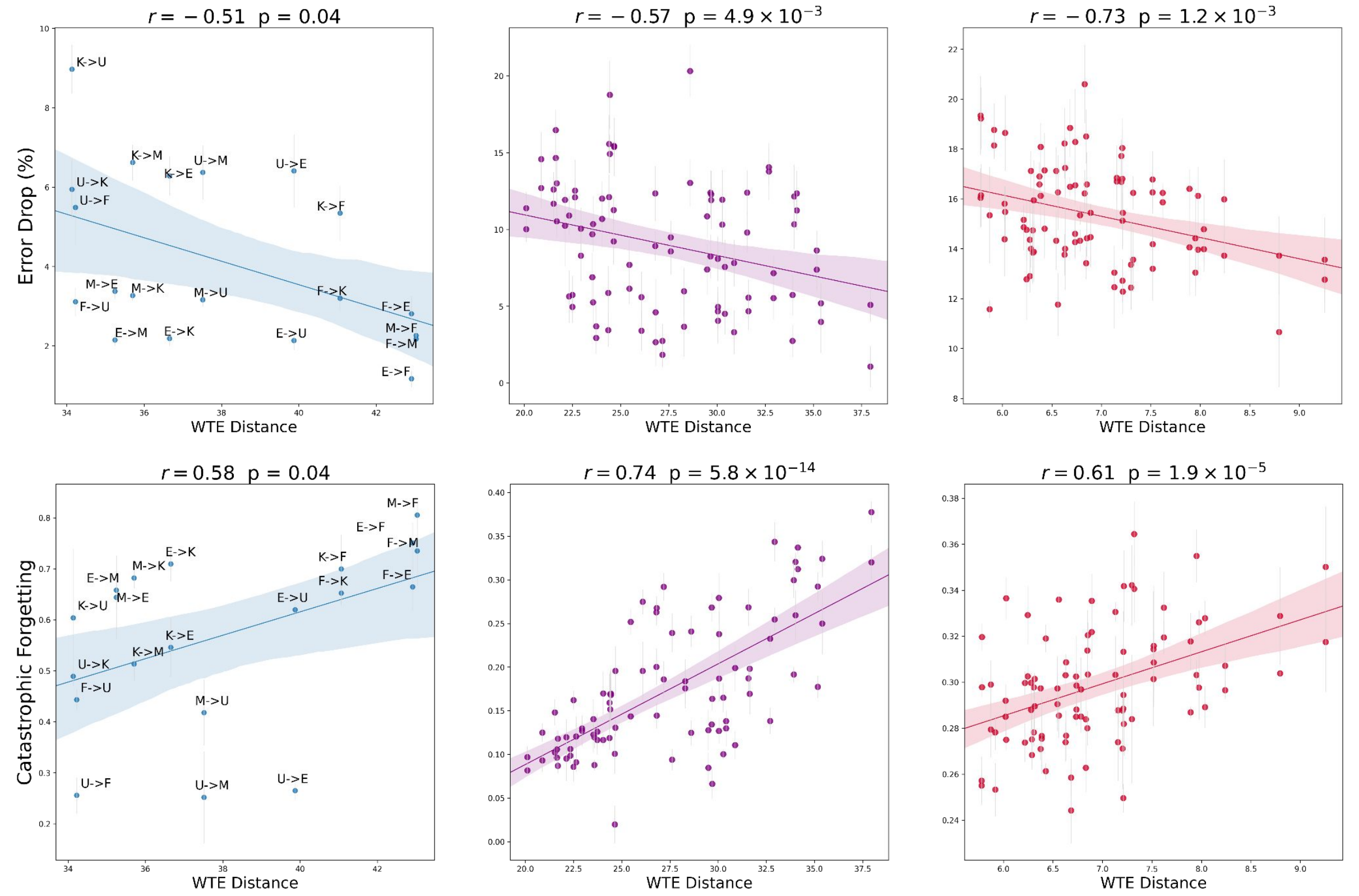}
    \caption{(Top row) forward transfer error drop and (bottom row) catastrophic forgetting against WTE distance on *NIST (left), Split-CIFAR100 (middle) and Split-Tiny ImageNet (right) over five runs. Pearson's $r$ and the corresponding $p$-value are reported on top of each experiment setting.}
    \label{fig:results}
\end{figure*}

\textit{Wasserstein Embedding}.
By Eq. \ref{eq:lbl_emb} and Eq. \ref{eq:otdd}, OTDD can be approximated by the squared 2-Wasserstein distance between the distributions over input-(label MDS embedding) pairs, $\tau'$. Then we leverage the Wasserstein embedding framework to embed the updated task distributions into a Hilbert space, with the goal of reducing the cost of computing pairwise Wasserstein distances. Again, we emphasize that this can bring down the cost from quadratic to linear with the number of task distributions. 

The WTE algorithm is summarized in Algo. \ref{alg:wte}. The outputs are the vector representations of input tasks with respect to a pre-determined MDS dimension and WE reference distribution.


\begin{algorithm}
    \caption{Wasserstein Task Embedding}
        \label{alg:wte}
    \begin{algorithmic}
    \Procedure{WTE}{$\{X_i = \{(x^i_n, y^i_n)\}_{n=1}^{N_i}\}_{i=1}^I$}
    \State Calculate label-to-label distance matrix W (Eq. \ref{eq:bures})
    \State Calculate $\psi(\nu_y)$ for all distinct labels $y$ (Algo. \ref{algo:mds})
    \State Stack each input with its label vector: $x\rightarrow[x, \psi(\nu_y)]$
    \State Calculate the WE $v_i=\Phi(\{[x^i_n, \psi(\nu_{y^i_n})]\}_{n=1}^{N_i})$ (Eq. \ref{eq:disrete_we})
    \Return $\{v_i\}_{i=1}^I$
    \EndProcedure
    \end{algorithmic}
\end{algorithm}

\section{Experiments}
To assess the effectiveness of our WTE framework, we empirically validate the correlation between WTE distance and forward/backward transferability on several datasets. Moreover, we provide both qualitative and quantitative comparison results with OTDD, and show WTE distance is well aligned with OTDD, and meanwhile is notably faster to compute. We use the MDS toolkit in scikit-learn and the exact linear programming solver in Python Optimal Transport (POT) \cite{flamary2021pot} library for implementing WE. We carry out the distance calculations on CPU and all the model training experiments on a 24GB NVIDIA RTX A5000 GPU.
\begin{figure*}[t!]
    \centering
    \includegraphics[width=0.8\linewidth]{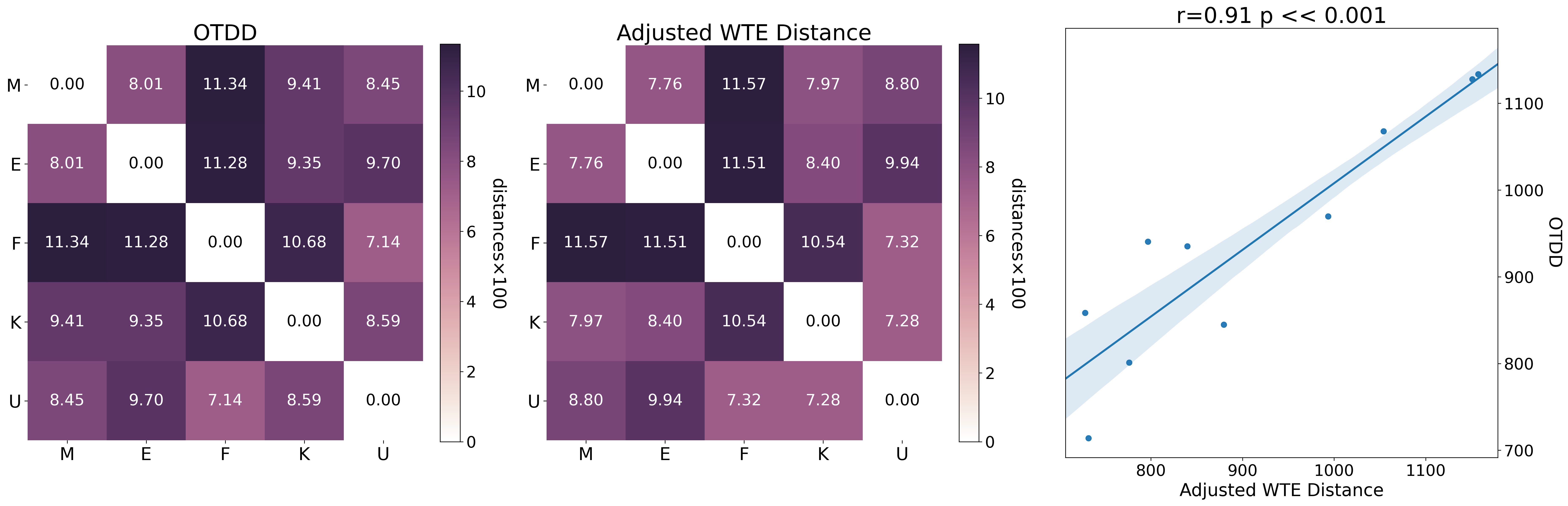}
    \vspace{-0.3cm}
    \caption{Pairwise OTDD (left) and WTE distances (middle) on the *NIST task group, and their correlation diagram (right). Notice that OTDD (Eq. \ref{eq:otdd}) is the squared $\mathcal{W}_2$, we report the squared WTE distances and adjust to the same scale according to the cost function. Adjusted WTE distance is strongly correlated with OTDD, with correlation coefficient $r=0.91$.}
    \label{fig:approx_otdd}
\end{figure*}
\subsection{Datasets}
We conduct experiments on the following three task groups:

\textit{*NIST} task group consists of the handwritten digits dataset MNIST \cite{LeCun2005TheMD} and its extensions EMNIST \cite{cohen2017emnist}, FashionMNIST \cite{xiao2017fashion}, KMNIST \cite{Clanuwat2018DeepLF} along with USPS \cite{291440}. We choose the mnist split for EMNIST dataset and thus all tasks contain 10 classes of gray-scale images. All datasets have a training set of 60,000 samples and a test set of 10,000 samples, except USPS, with a total of 9,298 samples. We resize the images from USPS into $28\times28$ pixel level to match with the others.

\textit{Split-CIFAR100} task group is generated by randomly splitting the CIFAR-100 \cite{Krizhevsky2009LearningML} dataset with 100 image categories into 10 smaller tasks, each of which is a classification with 10 classes. There are 600 $32\times32$ color images in the training set and 100 in the test set per class.

\textit{Split-Tiny ImageNet} task group follows the same splitting scheme as in Split-CIFAR100. We randomly divide the Tiny ImageNet \cite{Le2015TinyIV} into 10 disjoint tasks with 20 classes. Each class contains 500 training images, 50 validating images and 50 test images. For better model performance, we first rescale each sample to $256\times256$ and then perform a center crop to get $224\times224$ pixel images.

\subsection{Results}
To study the transfer behaviors against the WTE distances, we fix a model architecture for each task group. Specifically, we use ResNet18 \cite{he2016deep} on both *NIST and Split-Tiny ImageNet, and ResNet34 \cite{he2016deep} on Split-CIFAR100. In the forward transfer setting, for each source-target task pair, we first train the head (i.e., the classifier) of a randomly initialized backbone on the target task, and use the test performance as the baseline. Next, we adapt from a model pre-trained on the source task and finetune the head on the target task. We define the forward transferability of the source-target pair as the performance gain, i.e. error drop when adapting from the source task. To analyse backward transfer, all source tasks are trained jointly during the first phase to avoid task bias, then in the second phase the model learns only the target task and suffers from ``forgetting" the previous tasks. We use the catastrophic forgetting, i.e., negative backward transfer as a measure of backward (in)transferability. In implementations of WE, the reference distribution is fixed for each task group, and is randomly generated by upsampling random images at a lower spatial resolution to entail some smooth structure.

\begin{figure}
    \centering
    \includegraphics[width=\linewidth]{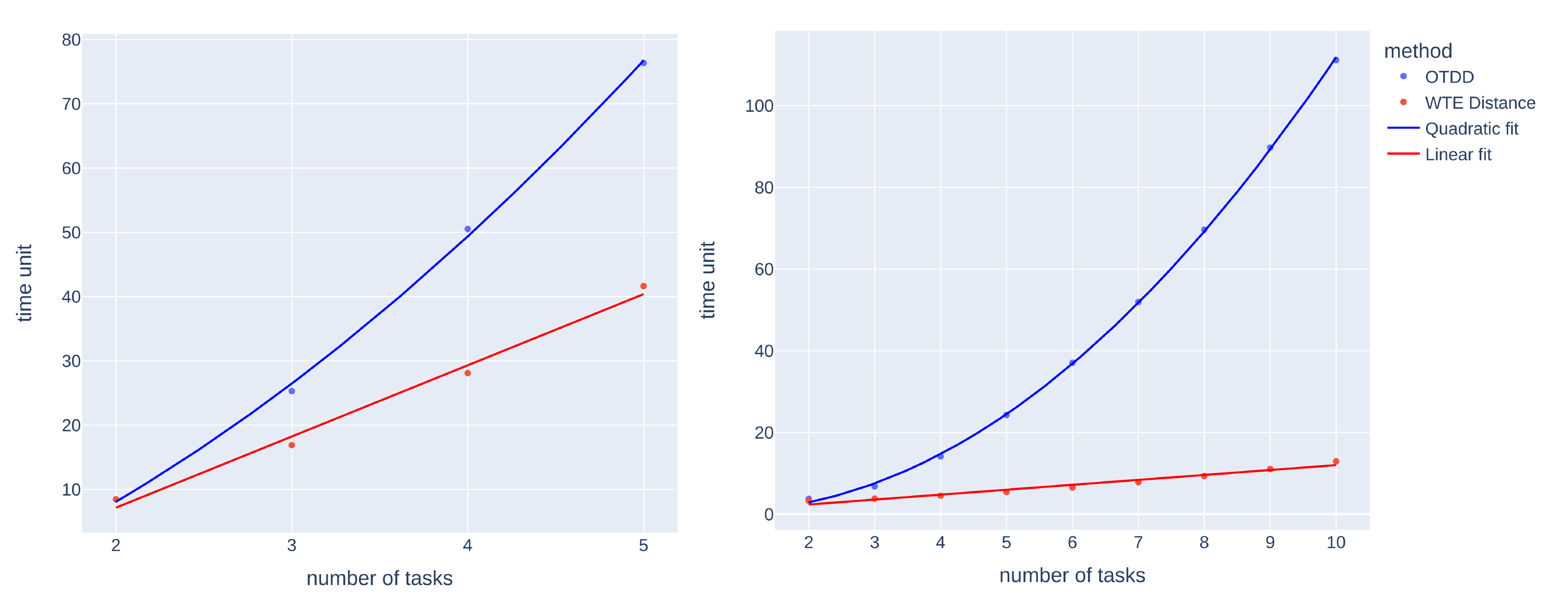}
    \caption{Wall-clock computation time comparison on the *NIST (left) and Split-CIFAR100 (right) task groups.}
    \label{fig:time}
\end{figure}
Fig. \ref{fig:results} summarizes the correlation diagrams between our proposed WTE distance and the forward/backward transferability on the aforementioned three task groups. WTE distance is negatively correlated with forward transferability, and positively correlated with catastrophic forgetting. In all scenarios, the correlation is strong and statistically significant, which confirms the efficacy of WTE distance as a measure of task similarities. We also visualize the comparison between WTE distance and OTDD on the *NIST task group in Fig. \ref{fig:approx_otdd}, showing strong correlation between the two distances.  
\subsection{Computation Complexity}
As we mentioned before, OTDD suffers from a prohibitive computational cost as the number of tasks grows large. The pairwise OTDD calculation for a set of $M$ tasks requires $\mathcal{O}(M^2N^3\log(N))$ time in the worst case, where $N$ is the largest number of samples among the tasks. WTE distance requires solving only $M$ optimal transport problem, leading to $\mathcal{O}(MN^3\log(N))$ complexity. To better demonstrate the efficiency of WTE distance, we report the wall-clock time comparison on the *NIST and Split-CIFAR100 in Fig. \ref{fig:time}.

\section{Conclusion}
In this paper, we propose Wasserstein task embedding (WTE), a model-agnostic task embedding framework for measuring task (dis)similarities in supervised classification problems. We perform a label embedding through multi-dimensional scaling and leverage the 2-Wasserstein embedding framework to embed tasks into a vector space, in which the Euclidean distance between the embedded points approximates the 2-Wasserstein distance between tasks. We demonstrate that our proposed task embedding distance is correlated with forward and backward transfer on *NIST, Split-CIFAR100 and Split-Tiny ImageNet task groups while being significantly faster than existing methods. In particular, we show statistically significant negative correlation between the WTE distances and the forward transfer, and positive correlation with the catastrophic forgetting (i.e. negative backward transfer). Lastly, we show the alignment of WTE distance with OTDD, but with a significant computational advantage as the number of tasks grows.

\bibliography{aaai23.bib}
\end{document}